# Improved Dynamic Time Warping (DTW) Approach for Online Signature Verification

**Azhar Ahmad Jaini[1], Ghazali Sulong[1] and Amjad Rehman[1]**

[1]Faculty of Computing Universiti Teknologi Malaysia, 81310 Skudai Johor Malaysia

**Abstract**—Online signature verification is the process of verifying time series signature data which is generally obtained from the tablet-based device. Unlike offline signature images, the online signature image data consists of points that are arranged in a sequence of time. The aim of this research is to develop an improved approach to map the strokes in both test and reference signatures. Current methods make use of the Dynamic Time Warping (DTW) algorithm and its variant to segment them before comparing each of its data dimension. This paper presents a modified DTW algorithm with the proposed Lost Box Recovery Algorithm aims to improve the mapping performance for online signature verification.

## 1. INTRODUCTION

Signing a letter, check, receipts with a unique signature is a common practice of verification of documents and are accepted as identity verification [1-10]. Modern age still accepts offline signature is a valid mode of authentication on documents. The legal system also requires handwritten signatures on contracts or legal documents to validate the documents and handwritten signature has been accepted widely accepted as valid identity verification[11-15]. However, with the rapid modernization of the world, verification of identity has become extra challenging and signatures are no longer considered safe identity verifications. Nowadays, there are a lot of fraud cases involving fraud signatures on cheques and credit cards [16,25]. That leads to a tighter identity verification system. A reliable identity verification system must ensure that only the identity owner could present such proof of identity. Besides signatures, fingerprints are also widely accepted and have been used for quite some time. However, with the evolution of biometrics technology, there are many other types of identity verifications. These newly found type of identity verifications are believed safer than older type of biometrics identification due to the nature of the older ones. The fingerprint, for example, is a form of a stamp from the physical texture of fingers which could be easily imitated by reconstructing synthetic fingers with fingerprints [26-30].

There are also several types of identities in the field of biometrics [31-35]. Among others are face, fingerprint, palm, iris, DNA, signature and voice. The more modern identities like iris and DNA are considered a lot safer than the older ones. However, due to the reasons discussed earlier, signatures are still socially acceptable [36-40]. So, due to this social behavior, the challenge is not only about changing to a new method of identity verification but to make the currently accepted way of identity verification as safe as possible as well. The motivation to make the signature as a safe identity verifier has brought many researchers to this field of study. Nowadays, a new approach to applying signature has been developed [41-45]. Online signature is one example of a modernized way of applying the signature to make it safer yet maintaining it as a relevant identity verifier in these modern days [46-50]. The further paper is organized into three main sections, section 2 presents the proposed methodology, section 3 exhibits simulation results and section 4 draws a conclusion.

## 2. PROPOSED METHODOLOGY

The data used in this project are from a reliable source that was being used in the Signature Verification Competition 2004 (SVC2004) which consist of 3200 signatures of genuine and skilled forgeries. This data is still being used by recent researchers as a benchmark. The Signature Verification section receives signature data from the earlier section as either reference data or test data. The data received must be of the standard format used by the verifier. This project uses a common online signature format. This very same format was used by almost all researchers in this domain.

The proposed Signature Verification Workflow is described in figure 1. Some of the data was not used in this project. Data like pen pressure and pen angle is among the data that will not be processed in this study. This study is working only with the first four columns namely, time, x-position, y-position, and pen condition (pen on/off, up/down). Although the rest of the data are not deleted, the unwanted columns are not going to be processed at all in the proposed workflow. This is because this study solely depending on pen location and time to reverse engineer the time series data back to stroke skeletons.

### 2.1 EXTREMES SEARCH

The first step is to detect the extremes of the data. This is the point where x or y is at their local extreme. This is done by comparing each point with its adjacent points. However, in this study, the common differential method of three points was not used. This is in consideration that such a detection method is too sensitive and all zero gradients will be detected as extremes. So, to solve this issue, a method to compare the value of the point with a number of adjacent points is proposed.

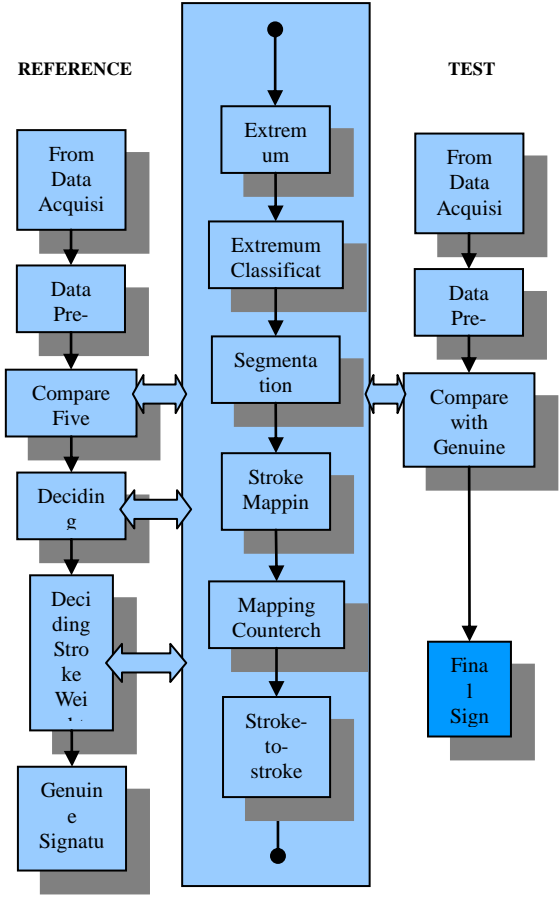

Figure 1: Proposed Signature Verification Workflow

A comparison of *l* points before and after the tested point is a set of points, T as described below.

$$T_x = \{x(t-l), x(t-l+1), ..... x(t+l-1), x(t-l)\}$$
$$T_y = \{y(t-l), y(t-l+1), ..... y(t+l-1), y(t-l)\} \quad (1)$$

Where t is sequential numbers representing each time interval. This will detect a series of extremes along the data. Then a point is considered as local maxima, xmax or local minima, xmin if it complies with their respective rules as stated below.

For x-maximum,
$$\exists! x_{max} \in T_x : x_{max} = max(T_x) \quad (2)$$
For x-minimum,
$$\exists! x_{min} \in T_x : x_{min} = min(T_x) \quad (3)$$

Furthermore, the existence of ripples needs to be considered. To avoid ripples, for any two extremes, x(tk) and x(tl), they cannot be too close together and to avoid that a mandatory time gap between the two extremes, S is proposed.

$$|t_k - t_l| > S \quad (4)$$

Where S is a pre-determined threshold.
Whereas, for y-maximum,
$$\exists! y_{max} \in T_y : y_{max} = max(T_y) \quad (5)$$

For y-minimum,

$$\exists! y_{min} \in T_y : y_{min} = min(T_y) \quad (6)$$

Just like in x extremes, the same mandatory gap in the y-extremes is also proposed to avoid vertical ripples. So, for any two extremes, y(tk) and y(tl), there shall exist a gap of at least *S* points that separating the two.

The reasons of putting *l* and *S* here are to eliminate the rippling effect discussed above. *l* acts as follows:
i) Making sure that the extremes point selected is not just a local extreme to their adjacent points but spread to a wider set of points, which is *l* points before and after the tested point.
ii) To make sure that the tested point is the only extreme throughout *l* points before and after the tested point. Whereas *S*, on the other hand, is to make sure, even if extremes points is detected, it still cannot be too close to each other. In this case, if a point is detected as extreme, the next extreme of the same class could only be declared after *S* points.

## 2.2 EXTREMES CLASSIFICATION

After all extremes points are detected, the next step is to classify the points. Here is where the full classification of the extremes takes place. This study proposed to classify extremes into nineteen classes. In addition to first, 4-bit based 16 classes, three classes are proposed.

In the proposed naming system, the four bits represent the following: The first two bits represent the quartet of the origin of the stroke. The third bit represents the type of the stroke either it is a second quadrant (basic type) or a single quartet (combinatorial type). The fourth bit represents the rotation of the stroke either clockwise or anti-clockwise.

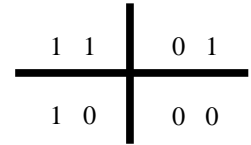

Figure 2: Quartet naming system

There are two types of strokes in this study, namely two-quartet type or basic type and a single-quartet type or combinatorial type. A two-quarter stroke is a basic type of strokes. The stroke starts at one quartet, goes to the centre and ends at the adjacent quartet whereas, the single-quartet is for strokes that starts and ends at the same quartet after going through the middle point in between the strokes. Figure 3 shows examples of common strokes. This stroke type is represented by the third bit where the two-quartet type is represented by bit '0' and the single-quartet type is represented by bit '1'.
Other than a quartet, the rotation of the stroke is important as well. Clockwise strokes in this study are represented by bit '1' and counterclockwise strokes are represented by bit '0'. As both strokes in figure 3 are clockwise type, they are both represented by bit '1'
So, for the stroke shown in the example in figure 3, the stroke is originating from bottom left quartet, which is quartet '10'. It is a two-quartet type represented by bit '0'. Then the rotation of the stroke is clockwise, represented by bit '1'. So the bit representation of the stroke in figure 3(a) is '1001' or '9' in its decimal form and stroke in figure 3(b) is represented by '1011'.
The three additional classes of strokes are as the follows.

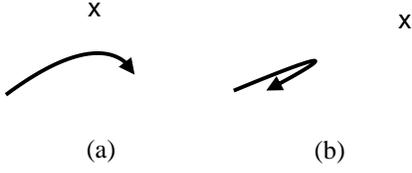

Figure 3: Stroke sample

i. Type 16: A touchdown – When pen starts to write
ii. Type 17: A takeoff – When pen starts to leave the media
iii. Type 18: A dot – A pen 'on' beside two pens 'off'

The type of strokes is decided by its immediate neighbors. This project only considers one point before and after the point that is going to be tested. There are cases where the immediate neighbor is not located in any quartet but in between the two quartets. For cases like this, the opposite neighbor will decide which quartet the point belongs to. If for example, the point, p(t-1) is in the middle of the quartet axis, it will be rotated toward p(t+1) via the shortest path with p(t) as center point. This could be seen clearer in figure 3(b). By doing this the peak type could be decided and mapping of similar strokes is possible.

Finally, the last attribute of a stroke is the rotation of the strokes. Rotation of a stroke could be checked by a simple cross product, C of the two vectors. This is formulated as below,

$$C = P_1 \times P_2$$

Where,
$$P_1 = (x_2 - x_1, y_2 - y_1)$$
$$P_2 = (x_3 - x_2, y_3 - y_2)$$
$$P_1 . P_2 = \begin{cases} clockwise, & C \geq 0 \\ anticlockwise, & C < 0 \end{cases} \quad (7)$$

## 2.3 CHARACTER SEGMENTATION

Character segmentation is a crucial step in the signature verification process [25,26]. Following, all extremes have been detected and classified, the next step is to segment them into strokes. Stroke is a series of points from an extreme to another extreme. Logically, if a stroke is formed between two nearby extremes, the curve is becoming too short that it would promote false acceptance. On the other hand, if a stroke is formed by too many extremes, it is contradicting with the motive of earlier segmentation process which is segmenting longer curves to become shorter for the ease of comparison process.

In this project, stroke is defined as the curve or line between extremes that are not too close to each other since stroke that is too short promotes false acceptance due to incapability to differentiate short strokes of either genuine or forgery. It shall also not too long that may promote false rejection due to incapability to compare and accept two genuine strokes. To construct an image data, the online signature needs to be reversed, engineer. This is done by connecting all the point in the time series data following the sequence of time.

There are many ways of connecting the points. One logical way is by applying the spline curve. This will surely eliminate jagged lines. However, in this study, points are connected with straight lines. The reasons for that are,

i. Points are too close to each other. Connecting them with straight lines is adequate to construct smooth curves.
ii. Spline smoothen sharp peaks of the signature whereby peaks needs to be as sharp as its original form.

So by doing this kind of segmentation, each stroke has two maximum-y and at least one minimum-y. There may be additional extremes in between, for instance, when there is takeoff or touchdown in the middle of the two maximum-y.

## 2.4 STROKE MAPPING

The DTW is used to map the strokes of a signature with similar strokes from other signature. As proposed by many earlier researchers, this project uses DTW algorithm. The DTW being applied here is based on one-to-one matching. However, since an extreme maybe matched with a few combinatorial extremes as long as it suits the combinatorial rules, the traditional one to one matching is becoming an issue. To solve this, a Multiple Layer DTW is proposed. In this Multiple Layer DTW, the DTW path is not only going to travel in the same matrix but the path could also hop to another layer of similar coordinate that are having a better matching value.

Basically, the similarity maximization technique is still obtained. However, the proposed modified version of Similarity Maximization S(i,j) is defined as,

$$S(i,j) = s_{i,j} + MAX \begin{Bmatrix} \overset{K}{\underset{k=i+1}{M}} S_1(k, j+1), \overset{L}{\underset{l=j+1}{M}} S_1(i+1,l), \\ \overset{M}{\underset{m=i+1}{M}} S_2(m, j+1), \overset{N}{\underset{n=j+1}{M}} S_2(i+1,n), \\ \overset{P}{\underset{p=i+1}{M}} S_3(p, j+1), \overset{Q}{\underset{q=j+1}{M}} S_3(i+1,q) \end{Bmatrix}$$
(8)

Where S(i,j) is the newly proposed modified Similarity Maximization, $S_n$ is the score of each pair at *n*-th layer of DTW matrix. The new path will look like the path shown in figure 4(d).

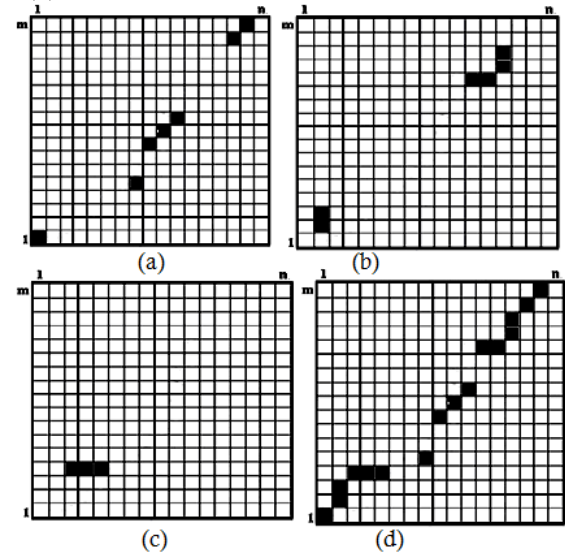

Figure 4(a) DTW path of first layer (b) DTW path of second layer (the combinatorial layer), (c).DTW path of third layer (the combinatorial layer) (d) The final DTW path

## 2.5 LOST BOX RECOVERY ALGORITHM

The DTW path generated by the Similarity Maximization Technique that has been used by earlier researchers even after the proposed Modified Similarity Maximization may not be as accurate as is hoped. For that, a new approach to supporting the decision made by the Modified Similarity Maximization Technique is proposed.

For every mapping decision done using Similarity Maximization Technique, a counter-checking algorithm is proposed to make sure that the decided path is the best possible path. This algorithm, namely the Lost Box Recovery Algorithm, will counter-check every node in the Lost Boxes and calculate the new path score if the path is to run through the node. In any case where the DTW Path via the tested node is having better score than the decided node by the Similarity Maximization, the decided node will be considered as a wrong path and the nodes will be deleted. Decision of the new path will be made again by the Similarity Maximization Techniques.

This is better elaborated by the following formula where decision, V is the decision switch either or not to apply the Lost Box Recovery Algorithm.

$$V = S_{k,l} - MAX\left\{ \underset{y=j}{\overset{l}{MAX}}\left(\underset{x=i}{\overset{m}{MAX}}\left(S_{x,y}\right)\right), \underset{v=i}{\overset{k}{MAX}}\left(\underset{w=j}{\overset{n}{MAX}}\left(S_{v,w}\right)\right)\right\} \quad (9)$$

where,

$(i, j)$ is the current node in $\Pi$

$(k,l)$ is the decided next node in $\Pi$

$(m,n)$ is the size of $\Pi$

$S_{(a,b)}$ is the score at node (a,b) using formula (21)

The value of V is then used to decide the value of new $S_{k,l}$ where,

$$\left(S_{k,l}, a, b\right) = \begin{cases} (S_{k,l}, k, l), & V \geq 0 \\ (0, i, j), & V < 0 \end{cases} \quad (10)$$

where, $(a,b)$ is the position of DTW path after the Lost Box Recovery Algorithm.

The performance of the Lost Box Recovery Algorithm needs to be measured before any conclusion could be made. Since in online signature verification domain, earlier researchers had more interest to measure the final verification performance rather than the quality of DTW Path itself, no standard DTW Path performance indicator could be found. So, a method of measuring the DTW Path performance needs to be developed.

The main factor to be considered is the actual correct mapping of the signatures. The decision that either the stroke are mapped to the correct stroke, although looks simple, is an expert decision in their own domain. The computer is unlikely to have the ability to decide which mapping algorithm is better than the others unless the data provided also provides the correct mapping as advised by experts [51-60].

In view of this, due to lack of expert data, a qualitative analysis is more suitable than the quantitative measurement. However, in classifying the quality of the mapping, the numerical approach is still being used since the real qualitative analysis could only be done by an expert. Even with this lack of expert input and lack of proposed method from earlier researchers, the mapping performance needs to be measured or else, no improvement will be made in this portion of the domain. To measure the mapping performance, these below mentioned rules are proposed.

i. The mapping test is between two genuine signatures.
ii. The test signature will be divided into five approximated sections.
iii. A section fails when any of the mappings in the section fails.
iv. The error is counted based on failed section and not each mapping.
v. Only obvious errors are accepted as the less perceivable errors are only decidable by an expert.
vi. The decision must be biased towards the old technique rather than the newly proposed technique.

Figure 5 shows a signature sample and how the sectioning is done. It has 40% error as strokes in section 2 and section 3 is shifted towards the earlier strokes

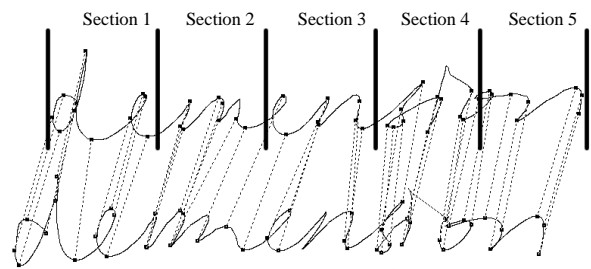

Figure 5: Signature is divided into section

The error value of the mapping could be simplified with the following formula.

$$Mapping\ Error = \frac{number\ of\ wrong\ sections}{5 \times Number\ of\ signatures} \quad (11)$$

## 3. EXPERIMENTAL RESULTS

From the said time series data, the signature could be reconstructed back as shown in figure 6. This is opposed to a common workflow which is normally starting their work by constructing x-position versus time or y-position versus time graphs.

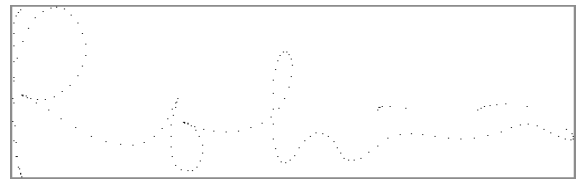

Figure 6: Signature points from the time series data

Here, all the points are joint with straight lines instead of the spline for the reasons stated earlier. From the figure, the lines connecting the points are smooth enough even printed at a relatively bigger size than the normal-sized signature. It could be seen in figure 7.

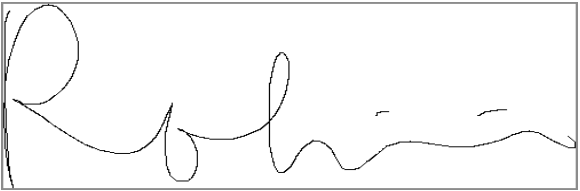
Figure 7: Reconstruct the signature – with lines

Figure 8 below shows the reconstructed signature is labeled with small boxes indicating extremes points. They are all extremes detected in the signature with additional proposed extremes namely touchdowns, take-offs, and points.

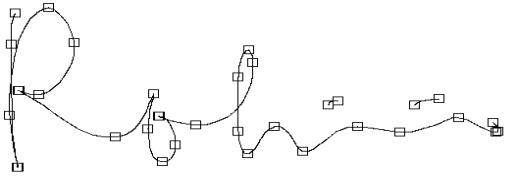
Figure 8: Detecting extremes

As what is discussed earlier, segmentation will not be done at all extremes points or else the stroke will be too short to compare thus promote false acceptance. Instead, in this study, segment borders are at y-maximum, touchdown, takeoff and points. The result could be seen in figure 9.

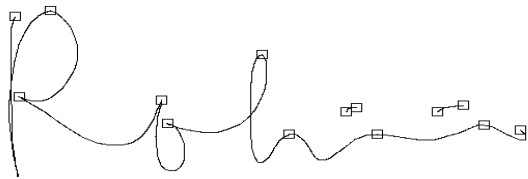
Figure 9: Segment at y-maximum

Figure 10 shows the classification of extremes that was detected in figure 8.

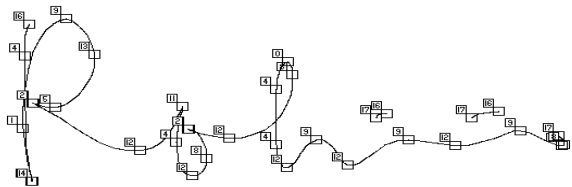
Figure 10: Type of extremes

After the extremes are detected using the proposed algorithm, the segments will go through a mapping process where the appropriate segments are mapped with another matching segment. Here, the discussions are based on sample 'user1_1.txt' and 'user1_2.txt'. To do this, the similarity matrix needs to be constructed first. The unprocessed Similarity Matrix could be seen in attachment 2. This is followed by searching the best path as mentioned in earlier chapters. The result of the maximum similarity matrix could be seen in attachment 3. To easily see this, figure 11 shows the signatures with mapping lines in between them.

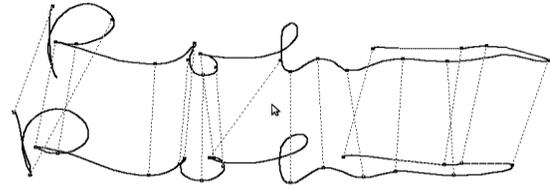
Figure 11: DTW matching using Similarity Maximization

In figure 11, there are a few incorrect matchings. This is one of the drawbacks in segment-to-segment that may yield an inconsistent result. This may not be realized in peak to peak matching but as the signature image is reconstructed, the incorrect matching's are clearly seen.

So, to overcome this, a countercheck measure on every hop in DTW path by checking back the decision is proposed. If the alternatives path has a better score, the earlier decided path will be deleted and the path search will be redone one step backward. This is done until the full path is decided as depicted in Figure 12.

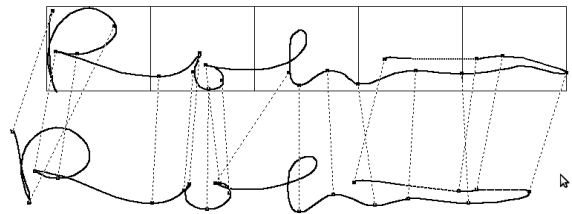
Figure 12: Divide signature into 5 sectors

The division into sectors here is an approximate division due to the decision of correct matching in signatures is an expert decision. This is mainly due to its dynamic behavior. For the signature above, for example, there might be more incorrect matching if this it is to be checked by authorized personnel on signature verification. However, there are at least two incorrect matching's that could clearly be seen. Each at sector 1 and sector 3. The above test for signature USER1 to USER5 and the result of the test is as stated in Table 1 below.

Table 1: Result of DTW Path Search Techniques

| Technique Name | Error Percentage |
| --- | --- |
| Similarity Maximization | 16.80% |
| Similarity Maximization with Lost Box Recovery | 3.60% |

### 4. CONCLUSION

Based on the experiment that has been done in this study, it is concluded that the mapping of the online signature using Similarity Maximization performed better with the proposed Lost Box Recovery Algorithm. There are several suggestions for the future works that might improve the overall performance of the algorithms used in this study. Among others, the extremes detection is good enough without the Gaussian Filter as Gaussian Filter will filter out the ripples along with the important information. However, if the Gaussian filter is used only to detect the location and the original ripples are maintained, this might help the exclusion of unwanted extremes. Besides that, the segmentation in this study divides the signatures into strokes. However the stroke in this study is loosely defined and for the purpose of this study, it is defined as the curves between y-extreme to another y-extreme. The fact that strokes in Latin, Chinese and Arabic signatures may be of different behavior might help on the overall performance of stroke-to-stroke verification.